\documentclass[10pt,twocolumn,letterpaper]{article}

\usepackage{cvpr}
\usepackage{times}
\usepackage{epsfig}
\usepackage{graphicx}
\usepackage{amsmath}
\usepackage{amssymb}

\usepackage{subfig}
\usepackage{float}

\usepackage[pagebackref=true,breaklinks=true,letterpaper=true,colorlinks,bookmarks=false]{hyperref}

\cvprfinalcopy 

\def\cvprPaperID{3941} 

\ifcvprfinal\fi

\usepackage{color}
\definecolor{green}{rgb}{0, 0.5, 0}
\definecolor{orange}{rgb}{0.8, 0.6, 0.2}
\definecolor{red}{rgb}{1.0, 0.0, 0.0}
\definecolor{teal}{rgb}{0.0, 0.4, 0.4}
\definecolor{purple}{rgb}{0.65,0,0.65}
\definecolor{saffron}{rgb}{0.95,0.75,0.2}
\definecolor{turquoise}{rgb}{0.0,0.5,0.5}

\usepackage[ruled,vlined,linesnumbered]{algorithm2e}
\usepackage{multirow}

\usepackage[normalem]{ulem}

\usepackage{mathtools}
\usepackage{amsmath, amsthm, amssymb, amsfonts, amsopn}
\usepackage{graphicx} 
\usepackage{textcomp}
\usepackage{xfrac}
\usepackage{bbm}

\usepackage{overpic}
\usepackage{subfig}
\usepackage{wrapfig}

\newcommand{\hidecomment}[1]{}
\newcommand{\R}{\mathbb{R}}

\newcommand{\AdjDec}{\mbox{\sc AdjDec}\xspace}

\newcommand{\SymDec}{\mbox{\sc SymDec}\xspace}

\newcommand{\BoxDec}{\mbox{\sc BoxDec}\xspace}

\begin{document}

\title{Im2Struct: Recovering 3D Shape Structure from a Single RGB Image}

\author{Chengjie Niu \quad\quad\quad Jun Li \quad\quad\quad Kai Xu\thanks{Corresponding author: kevin.kai.xu@gmail.com}\\
National University of Defense Technology\\
}

\maketitle
\thispagestyle{empty}


\begin{abstract}
We propose to recover 3D shape structures from single RGB images, where
structure refers to shape parts represented by cuboids and part relations encompassing connectivity and symmetry. Given a single 2D image with an object depicted, our goal is automatically recover a cuboid structure of the object parts as well as their mutual relations. We develop a convolutional-recursive auto-encoder comprised of structure parsing of a 2D image followed by structure recovering of a cuboid hierarchy.
The encoder is achieved by a multi-scale convolutional network trained with the task of shape contour estimation, thereby learning to discern object structures in various forms and scales.
The decoder fuses the features of the structure parsing network and the original image, and recursively decodes a hierarchy of cuboids.
Since the decoder network is learned to recover part relations including connectivity and symmetry explicitly,
the plausibility and generality of part structure recovery can be ensured.
The two networks are jointly trained using the training data of contour-mask and cuboid-structure pairs.
Such pairs are generated by rendering stock 3D CAD models coming with part segmentation.
Our method achieves unprecedentedly faithful and detailed recovery of diverse 3D part structures from
single-view 2D images. We demonstrate two applications of our method including structure-guided completion of 3D volumes reconstructed from single-view images and structure-aware interactive editing of 2D images.
%
\end{abstract}


\section{Introduction}

The last few years have witnessed a continued interest in single-view image-based 3D modeling~\cite{choy20163d,fan2016point,girdhar2016learning}.
The performance of this task has been dramatically boosted, due to the tremendous success
of deep convolutional neural networks (CNN) on image-based learning tasks~\cite{krizhevsky2012imagenet}.
The existing deep models, however,
have so far been mainly targeting the output of volumetric representation of 3D shapes~\cite{choy20163d}.
Such models are essentially learned to map an input 2D image
to a 3D image (voxel occupancy of a 3D shape in a 3D volume).
Some compelling results have been demonstrated.

\begin{figure}[!t]
\begin{center}
\includegraphics[width=0.96\linewidth]{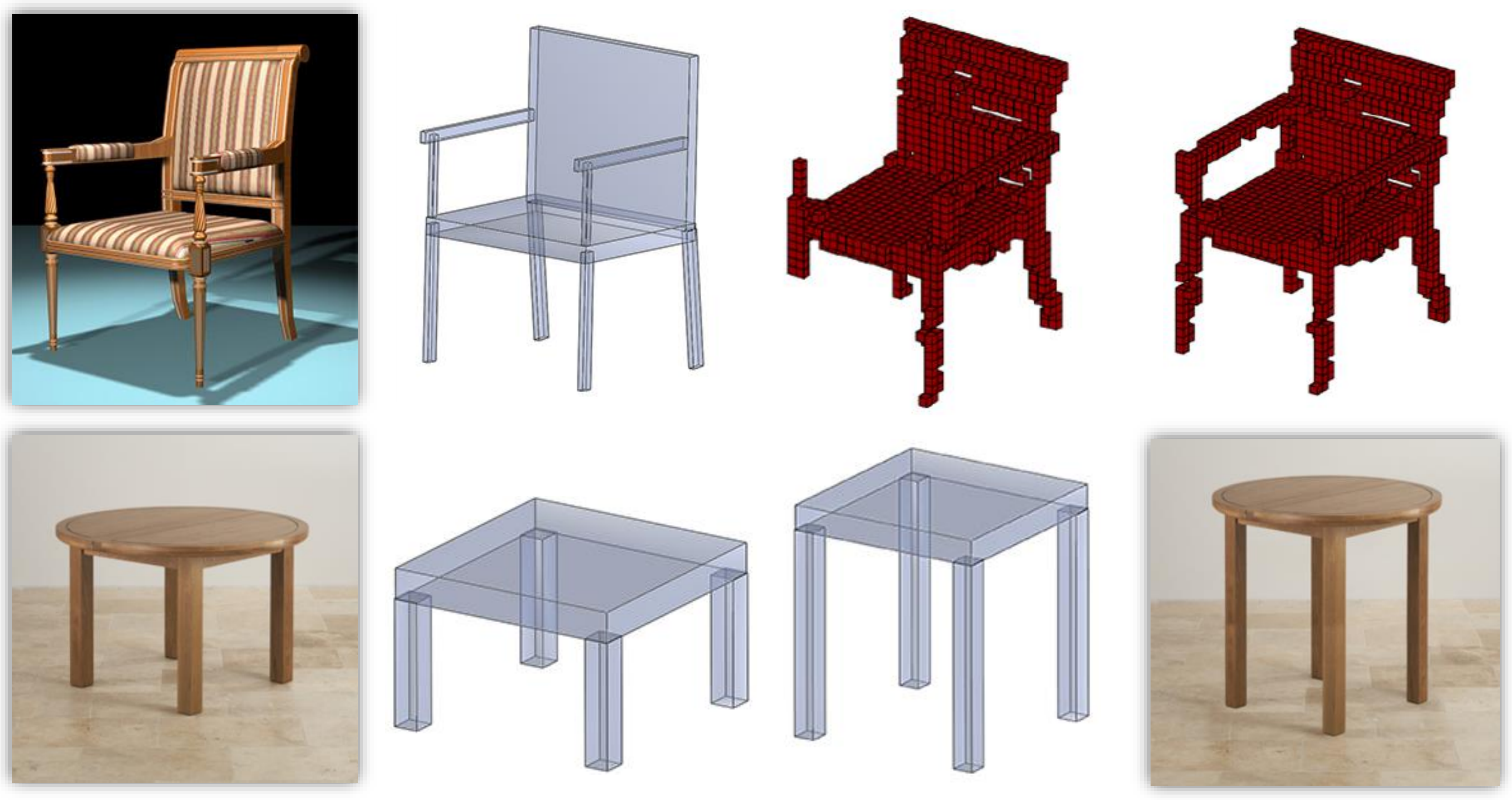}
\end{center}
   \caption{3D shape structures (2nd column) recovered from photos of household objects (1st column).
    Top row: The inferred 3D shape structure can be used to complete and refine the volumetric shape estimated from the image using existing methods~\cite{wu2016learning}. Bottom row: The structure is used to assist structure-aware image editing, where our cuboid structure is used as an editing proxy~\protect~\cite{zheng_sigg12}.}
\label{fig:teaser}
\end{figure}

\begin{figure*}[t]
\begin{center}
\includegraphics[width=0.99\linewidth]{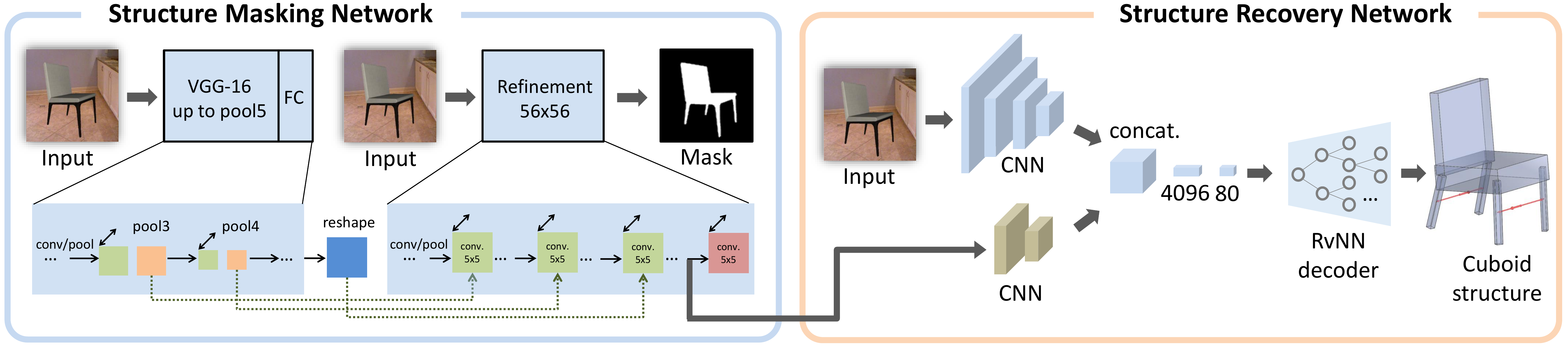}
\end{center}
   \caption{An overview of our network architecture. The structure masking network is a two-scale CNN which is trained to produce a contour mask for the object of interest. The structure recovery network first fuses the feature map of the masking network and the CNN feature of the original image, and decode the fused feature recursively into a box structure. The red arrows in the resultant chair structure (right most) indicate recovered reflectional symmetries between chair legs.
   }
\label{fig:overview}
\end{figure*}

While enjoying the high capacity of deep models in learning the image-to-image mapping,
the 3D volumes reconstructed by these methods lose
an important information of 3D shapes -- shape topology or part structure.
Once a 3D shape is converted into a volumetric representation,
it would be hard to recover its topology and structure, especially when there exist topological
defects in the reconstructed volume.
Shape structure, encompassing part composition and part relations, has been found highly
important to semantic 3D shape understanding and editing~\cite{mitra2013structure}.
Inferring a part segmentation for a 3D shape (surface or volumetric model) is known to be difficult~\cite{kalogerakis20163d}.
Even if a segmentation is given, it is still challenging to reason about part relations
such as connection, symmetry, parallelism, etc.

We advocate learning a deep neural network that directly recovers 3D shape structure
of an object, from a single RGB image.
The extracted structure can be used for enhancing the volumetric reconstruction
obtained by existing methods, facilitating structure-aware editing of the reconstructed 3D shapes,
and even enabling high-level editing of the input images (see Fig.~\ref{fig:teaser}).
However, directly mapping an image to a part structure seems a dunting task.
Tulsiani et al.~\cite{tulsiani2016learning} proposed a deep architecture to map a 3D volume to a set of cuboid primitives. Their method, however, cannot be adapted to our problem setting since the output primitive set
does not possess any structural information (mutual relations between primitives are not recovered).

Our problem involves the reasoning not only about
shape geometry, but also for higher level information of part
composition and relations. It poses several special challenges.
1) Different from shape geometry, part decomposition and relations
do not manifest explicitly in 2D images.
Mapping from pixels to part structure is highly ill-posed, as
compared to pixel-to-voxel mapping studied in many existing
2D-to-3D reconstruction works.
2) Many 3D CAD models of man-made objects contain diverse, fine-grained substructures. A faithful recovery of
those complicated 3D structures goes far beyond shape synthesis modulated by a shape classification.
3) Natural images always contain cluttered background and the imaged objects have large variations
of appearance due to different textures and lighting conditions.

Human brains do well both in shape inference based on low-level visual stimulus
and structural reasoning with the help of prior knowledge about 3D shape compositions.
The strength of human perception is to integrate the two ends of processing and reasoning
to form a capable vision system for high-level 3D shape understanding.
Motivated by this, we propose to learn and integrate two networks,
a \emph{structure masking network} for accentuating multi-scale object structures
in an input 2D image, followed by a \emph{structure recovery network} to
recursively recover a hierarchy of object parts abstracted by cuboids
(see Figure~\ref{fig:overview}).

The structure masking network produces a multi-scale attentional
mask for the object of interest, thereby decerning its shape structures in various forms and scales.
It designed as a multi-scale convolutional neural networks (CNN)
augmented with jump connections to retain shape details while
screening out the structure-irrelevant information such as background and textures in the output mask image.
The structure recovery network fuses
the features extracted in the structure masking network and
the CNN features of the original input image and feed them into
a recursive neural network (RvNN) for 3D structure decoding~\cite{li2017grass}.
The RvNN decoder, which is trained to explicitly model part relations,
expands the fused image features recursively into a tree organization of 3D cuboids
with plausible spatial configuration and reasonable mutual relations.


The two networks are jointly trained, with the training data of image-mask and cuboid-structure pairs.
Such pairs can be generated by rendering 3D CAD models and extracting the box structure based
on the given parts of the shape.
Several mechanisms are devised to avoid overfitting in training this model.
Experiments show that our method is able to faithfully recover diverse and detailed part structures of
3D objects from single 2D natural images. Our paper makes the following contributions:
\begin{itemize}
  \item We propose to directly recover 3D shape structures from single RGB images.
  The faithful and detailed recovery of 3D structural information of an object, such as part connectivity and symmetries, from 2D images has never been seen before, to our knowledge.
  \item We present an architecture to tackle the hard learning task, via integrating a convolutional structure masking network and a recursive structure recovery network.
  \item We develop two prototype applications where we use the recovered box structures 1) to refine
  the 3D shapes reconstructed from single images by existing methods and 2) to assist structure-aware
  editing of 2D images.
\end{itemize}



\section{Related work}

Reconstructing 3D shapes from a single image has been a long-standing pursue in both
vision and graphics fields.
Due to its ill-posedness, many priors and assumptions have been attempted,
until the proliferation of high-capacity deep neural networks.
We will focus only on those deep learning based models and categorize
the fast-growing literature in three different dimensions.

\paragraph{Depth estimation vs. 3D reconstruction.}
Depth estimation is perhaps the most straightforward solution
for recovering 3D information from single images.
Deep learning has been shown to be highly effective for depth estimation~\cite{eigen2015predicting,laina2016deeper}.
Compared to depth estimation, reconstructing a full 3D model is much more challenging due to
the requirement of reasoning about the unseen parts. The latter has to resort to shape or structure priors.
Using deep neural networks to encode shape priors of a specific category has received much attention lately,
under the background of fast growing large 3D shape repositories~\cite{Shapenet}.
Choy et al.~\cite{choy20163d} develop a 3D Recurrent Reconstruction
Neural Network to generate 3D shapes in volumetric representation, given a single image as input.
A point set generation network is proposed for generating from a 2D image a 3D shape in point cloud~\cite{fan2016point}.
We are not aware of any previous works that can generate part-based structures
directly from a single image.

\paragraph{Discriminative vs. Generative.}
For the task of 3D modeling from 2D images, discriminative models are learned to map an input
image directly to the output 3D representation, either by a deep CNN for one-shot generation
or a recurrent model for progressive generation~\cite{choy20163d}.
The advantages of such approach include ease of training and high-quality results.
With the recent development of deep generative models such as variational auto-encoder (VAE)~\cite{kingma2013auto},
generative adversarial nets (GAN)~\cite{goodfellow2014generative} and their variants.
Learning a generative model for 3D shape generation has gained extensive research~\cite{wu2016learning,girdhar2016learning,li2017grass}.
For generative models, the input image can be used to condition the sampling from the predefined parameter space
or learned latent space~\cite{wu2016learning,girdhar2016learning}.
Generative models are known hard to train. For the task of cross-modality mapping,
we opt to train a discriminative model with a moderate size of training data.

\paragraph{Geometry reconstruction vs. Structure recovery.}
The existing works based on deep learning models mostly utilize volumetric 3D shape representation~\cite{wu2016learning,girdhar2016learning}.
Some notable exceptions include generating shapes in point clouds~\cite{fan2016point}, cuboid primitives~\cite{tulsiani2016learning} and manifold surfaces~\cite{lun20173d}.
However, none of these representations contains structural information of parts and part relations.
Interestingly, structure recovery is only studied with non-deep-learning approaches~\cite{xu2011photo,su2014estimating,huang2015single}.
This is largely because of the lack of a structural 3D shape representation suitable for deep neural networks.
Recently, Li et al.~\cite{li2017grass} propose to use recursive neural networks
for structural representation learning, nicely addressing the encoding/decoding of arbitrary number of shape parts
and various types of part relations.
Our method takes the advantage of this and integrate it into a cross-modality
mapping architecture for structure recovery from an RGB image.


\section{Method}

We introduce our architecture for learning 3D shape structures from single images. It is an auto-encoder
composed of two sub-networks: a \emph{structure masking network} for decerning the object structures from the input 2D image and a \emph{structure recovery network} for recursive inference of a hierarchy of 3D boxes along with their mutual relations.
\subsection{Network architecture}
Our network is shown in Fig.~\ref{fig:overview}, which is composed of two modules:
a two-scale convolutional structure masking network and a recursive structure recovery network.
The structure masking network is trained to estimate the contour of the object of interest.
This is motivated by the observation that object contours provides strong cues for understanding
shape structures in 2D images~\cite{shotton2005contour,toshev2012shape}.
Instead of utilizing the extracted contour mask,
we feed the feature map of the last layer of the structure masking network into the structure recovery network.
To retain more information in the original image, this feature is fused with the CNN feature of
the input image via concatenation and fully connected layers, resulting in a $80$D feature code.
An RvNN decoder then recursively unfolds the feature code into a hierarchical organization of boxes,
with plausible spatial configuration and mutual relations,
as the recovered structure.
%

\subsection{Structure Masking Network.}
Our structure masking network is inspired by the recently proposed multi-scale network for detailed depth estimation~\cite{li2017ICCV}.
Given an input RGB image rescaled to $224 \times 224$,
we design a two-scale structure masking network to output a binary contour mask with a quarter of the input resolution ($56 \times 56$).
The first scale captures the information of the whole image while
the second produces a detailed mask map at a quarter of the input resolution.
As our prediction target is a binary mask, we use the SoftMax Loss as our training loss.

We employ VGG-16 to initialize the convolutional layers (up to pool5) of the first scale network, followed by two fully connected layers. The feature maps and outputs of the first scale network are fed into various layers of the second scale one for refined structure decerning. The second scale network, as a refinement block, starts from one $9 \times 9$ convolution and one pooling over the original input image, followed by nine successive $5\times 5$ convolutions without pooling. The feature maps from the pool3, pool4 and the output of last fully connected layer of first scale network are fused into the second, the fourth and the sixth convolutional layer of the second scale network, respectively. All the feature fusions get through a jump connections layer, which has a $5\times5$ convolutional layer and a 2x or 4x up-sampling to match the $56\times 56$ feature map size in the second scale; the jump connection from the fully connected layer is a simple concatenation.
It is shown that jump connections help extracting detailed structures from images effectively~\cite{li2017ICCV}.

\subsection{Structure Recovery Network}

The structure recovery network integrates the features extracted
from the structure masking network and for the input image into a bottleneck feature
and recursively decodes it into a hierarchy of part boxes.

\paragraph{Feature fusion.}
We fuse features from two convolutional channels. One channel takes as input the feature map of the structure masking network (the last feature map before the mask prediction layer), followed by two convolutions and poolings. Another channel is the CNN feature of the original image extracted by a VGG-16. The output feature maps from the two channels are then concatenated with size $7\times7$, and further encoded into a $80$D code after two fully connected layers, capturing the object structure information from the input image.
%
We found through experiments such fused features not only improve the accuracy of structure recovery, but also attain good domain-adaption from rendered images to real ones.
We believe the reason is that the extracted features for mask prediction task retain shape details through factoring them out of background clutters, texture variations and lighting conditions.
Since it is hard for the masking network to produce perfect mask prediction, the CNN feature of the original image provides complimentary information via retaining more object information.

\paragraph{Structure decoding.} We adopt a recursive neural network (RvNN) as box structure decoder like in~\cite{li2017grass}. Starting from a root feature code, RvNN recursively decodes its into a hierarchy of features until reaching the leaf nodes which each can be further decoded into a vector of box parameters.
There are three types of nodes in our hierarchy: leaf node, adjacency node and symmetry node.
During the decoding, two types of part relations are recovered as the class of internal nodes: \emph{adjacency} and \emph{symmetry}. Thus, each node can be decoded by one of the three decoders below, based on its type (adjacency node, symmetry node or box node):

\begin{description}
  \item[Adjacency decoder.] Decoder \AdjDec splits a parent code $p$ into two child codes $c_1$ and $c_2$, using the mapping function:
  \[
    [c_1 \ c_2] \ = \ \tanh(W_{ad} \cdot p + b_{ad})
  \]
  where $W_{ad} \in \R^{2n \times n}$ and $b_{ad} \in \R^{2n}$. $n=80$ is the dimension of a non-leaf node.

  \item[Symmetry decoder.] Decoder \SymDec recovers a symmetry group in the form of a symmetry generator (a node code $c$) and a vector of symmetry parameters $s$:
  \[
    [c \ s] \ = \ \tanh(W_{sd} \cdot p + b_{sd})
  \]
  where $W_{sd} \in \R^{(n + m) \times n}$, and $b_{sd} \in \R^{m + n}$. We use $m = 8$ for symmetry parameters consisting of: symmetry type ($1$D); number of repetitions for rotational and translational symmetries ($1$D); and the reflectional plane for reflective symmetry, rotation axis for rotational symmetry, or position and displacement for translational symmetry ($6$D).

  \item[Box decoder.] Decoder \BoxDec converts the code of a leaf node to a $12$D box parameters defining the center, axes and dimensions of a 3D oriented box, similar to~\cite{li2017grass}.
   \[
    [x] \ = \ \tanh(W_{ld} \cdot p + b_{ld})
  \]
   where $W_{ld} \in \R^{12 \times n}$, and $b_{ld} \in \R^{12}$.
\end{description}

\begin{figure}[!t]
\begin{center}
\includegraphics[width=1.0\linewidth]{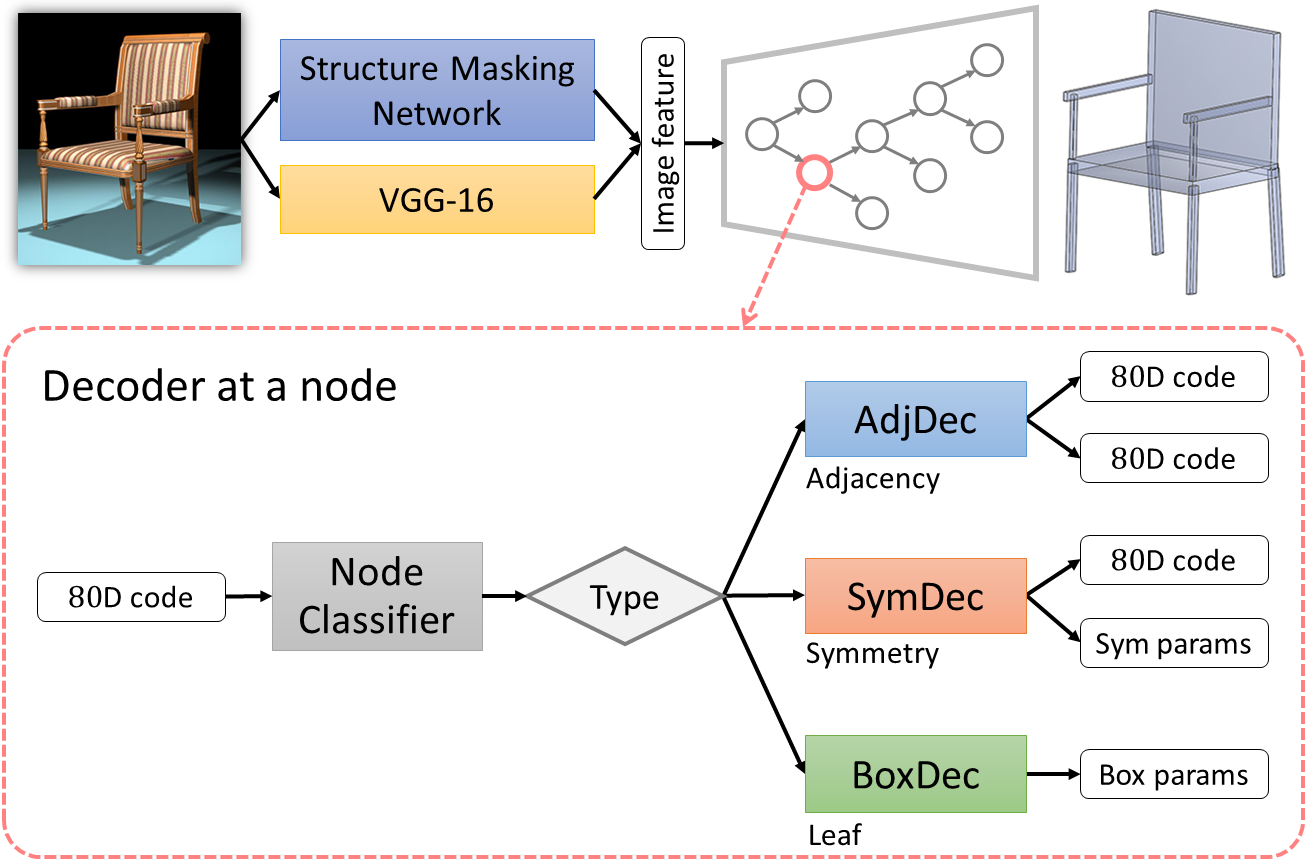}
\end{center}
   \caption{Recursive decoding of a 3D box structure from a 2D image feature (top) and an illustration of decoder network at a given node (bottom).}
\label{fig:decoding}
\end{figure}

The decoders are recursively applied during decoding.
The key is how to determine the type of a node so that the corresponding decoder can be used at the node.
This is achieved by learning a node classifier based on the training task of structure recovery where the ground-truth
box structure is known for a given training pair of image and shape structure.
The node classifier is jointly trained with the three decoders.
The process of structure decoding is illustrated in Fig.~\ref{fig:decoding}.
In our implementation, the node classifier and the decoders for both adjacency and symmetry are two-layer networks, with the hidden layer and output layer being $200$D and $80$D vectors, respectively.

\subsection{Training details}
\begin{figure}[b!]
    \centering
    \includegraphics[width=0.9\linewidth]{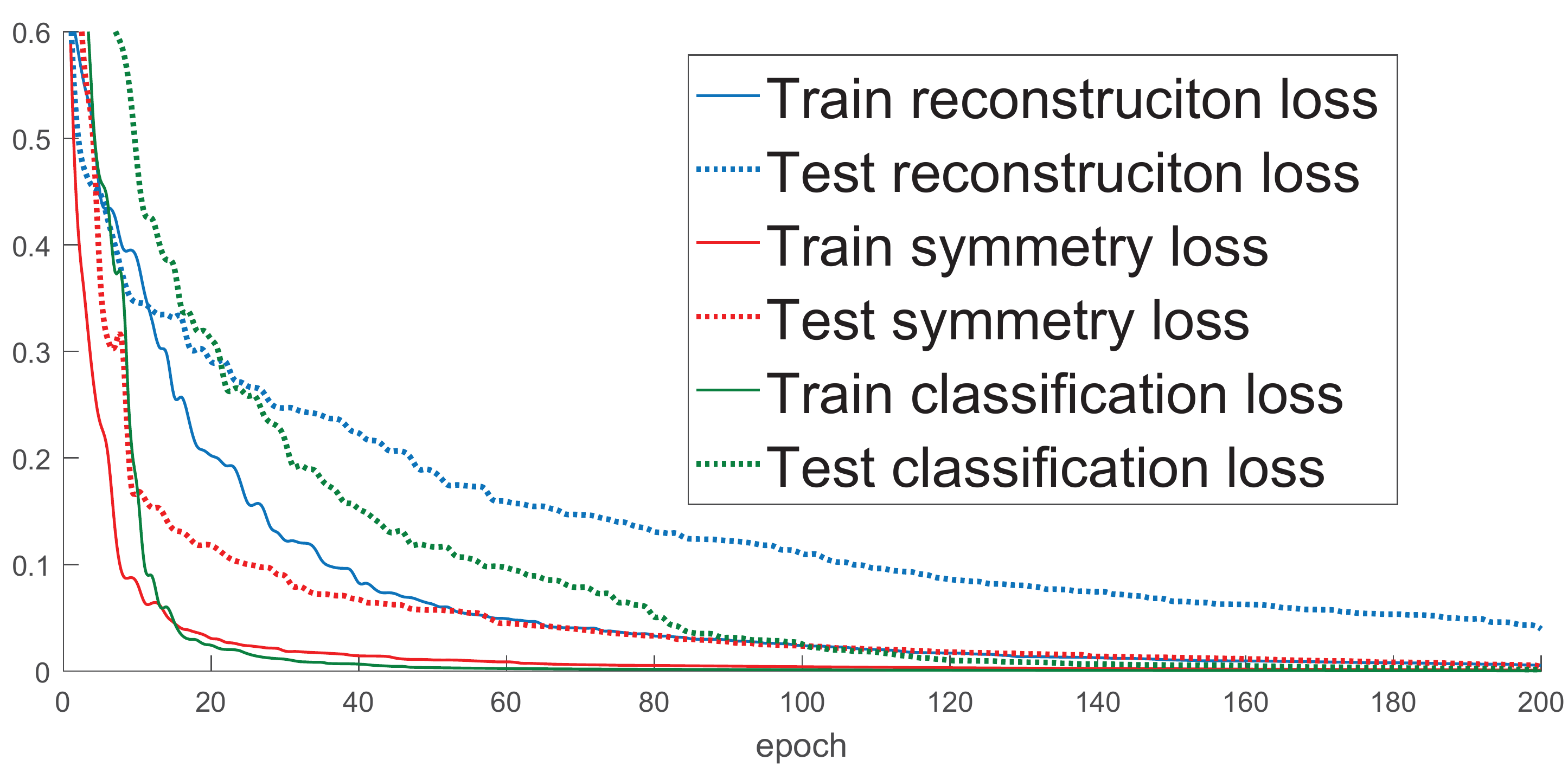}
    \includegraphics[width=0.9\linewidth]{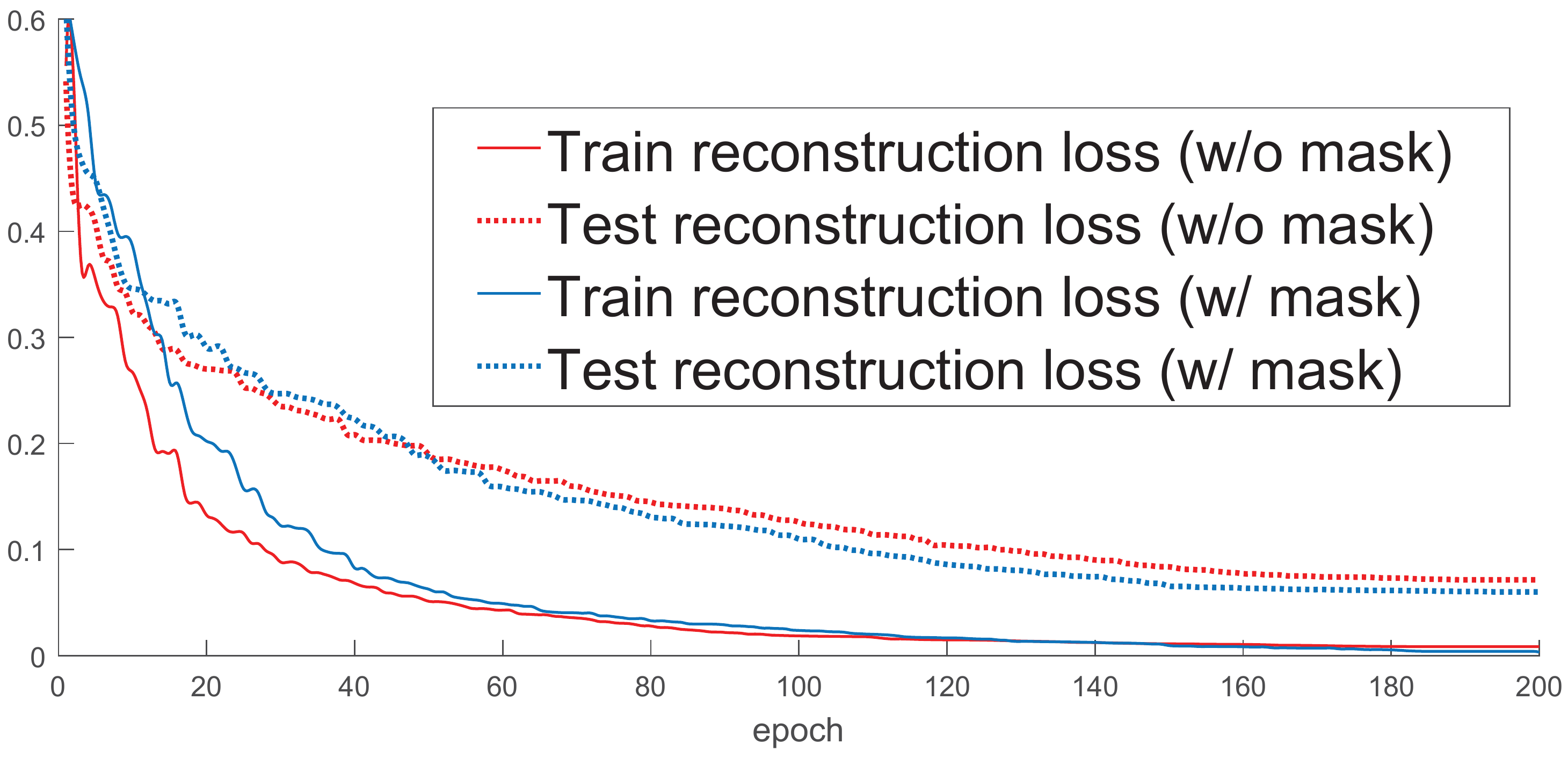}
    \caption{The convergence of the structure recovery network. We show in the top plot the training and testing losses for cuboid reconstruction, symmetry parameter recovery and node classification, respectively.
    The bottom plot shows that our method with structure masking network yields lower reconstruction loss than without it.}
    \label{fig:Loss}
\end{figure}

There are two stages in the training. First, we train the structure masking network to estimate a binary object mask for the input image. The first and the second scale of the structure masking network are trained jointly.
In the next, we jointly refine the structure masking network and train the structure recovery network, during which a low learning rate for structure masking network is used. The structure recovery loss is computed as the sum of the box reconstruction error and the cross entropy loss for node classification.
The reconstruction error is calculated as the sum of squared differences between the input and output parameters for each box and symmetry node. Prior to training, all 3D shapes are resized into a unit bounding box to make the reconstruction error comparable across different shapes. In Fig.~\ref{fig:Loss} (top), we plot the training and testing losses for box reconstruction, symmetry recovery and node classification, respectively, demonstrating the convergence of our structure recovery network.


We use the Stochastic Gradient Descent (SGD) to optimize our structure recovery network with back-propagation through structure (BPTT) for the RvNN decoder training. The convolutional layers of VGG-16 are initialized with the parameters pre-trained over ImageNet; all the other convolutional layers, the fully connected layers and the structure recovery network are randomly initialized. The learning rate of the structure masking network is $10^{-4}$ for pre-training and $10^{-5}$ for fine-tuning.
During joint training,
the learning rate is $10^{-3}$ for structure masking network, $0.2$ for RvNN decoder and $0.5$ for RvNN node classifier. These learning rates are decreased by a factor of $10$ for every $50$ epoches.
Our network is implemented with Matlab based on the MatConvNet toolbox~\cite{vedaldi2015matconvnet}.
The details on generating training data is provided in Section~\ref{train:data}.



\section{Experiments}

We collected a dataset containing $800$ 3D shapes from three categories in ShapeNet: chairs ($500$), tables ($200$), aeroplanes ($100$). The dataset is split into two subsets for training($70\%$) and testing ($30\%$), respectively.
With these 3D shapes, we generate training and testing pairs of image mask and shape structure to train the network and evaluate our method quantitatively. We also evaluate our methods qualitatively with a Google image search challenge. Both quantitative and qualitative evaluations demonstrate the capability of our method in recovering 3D shape structures from single RGB images faithfully and accurately.

\subsection{Training data generation}
\label{train:data}

\paragraph{Image-structure pair generation.}
For each 3D shape, we create $36$ rendered views around the shape for every $30^\circ$ rotation and with $3$ elevations.
Plus another $24$ randomly generated views, we create $60$ rendered RGB images in total for each shape.
The 3D shapes are rendered with randomly selected backgrounds from NYU v2 dataset.
For each RGB image, the ground-truth object mask can be easily extracted using the depth buffer for rendering.

All 3D shapes in our dataset are pre-segmented based on their original mesh components or using the symmetry-aware segmentation proposed in~\cite{wang2011}.
We utilize symmetry hierarchy~\cite{wang2011} to represent the shape structure, which defines how parts in a shape are recursively assembled by connectivity or grouped by symmetry. We adopt the method in~\cite{li2017grass} to infer consistent hierarchy trees for the shapes of each category.
Specifically, we train a unsupervised auto-encoder with the task of self-reconstruction for all shapes. During testing, we use this auto-encoder to perform a greedy search of grouping hierarchy for each shape.
For more details on this process, please refer to the original work.
Consequently, we generate $60$ image-structure pairs for each 3D shape.

\paragraph{Data processing and augmentation.}
To further enhance our dataset and alleviate overfitting, we conduct on each training 3D shape structure-aware deformation~\cite{xu2012fit} based on component-wise controllers~\cite{zheng2011component} to generate a set of structurally plausible variations for the training shape.
Such structure-aware deformation preserves the connection and symmetry relations between shape parts, while maintaining the shape texture for each part. This step is fully automatic and the parameters for each variation generation is randomly set within a given range. In our implementation, we randomly generate $20$ new variations for each 3D shape, thus enlarging our database to $16$K 3D shapes.
For the input images (and the corresponding object masks), we employ the common operations for image data augmentation~\cite{li2017ICCV} such as color perturbation, contrast adjustment, image flip and transformation, etc.

\subsection{Results and evaluation}

\begin{figure}[!t]
\begin{center}
\includegraphics[width=1.0\linewidth]{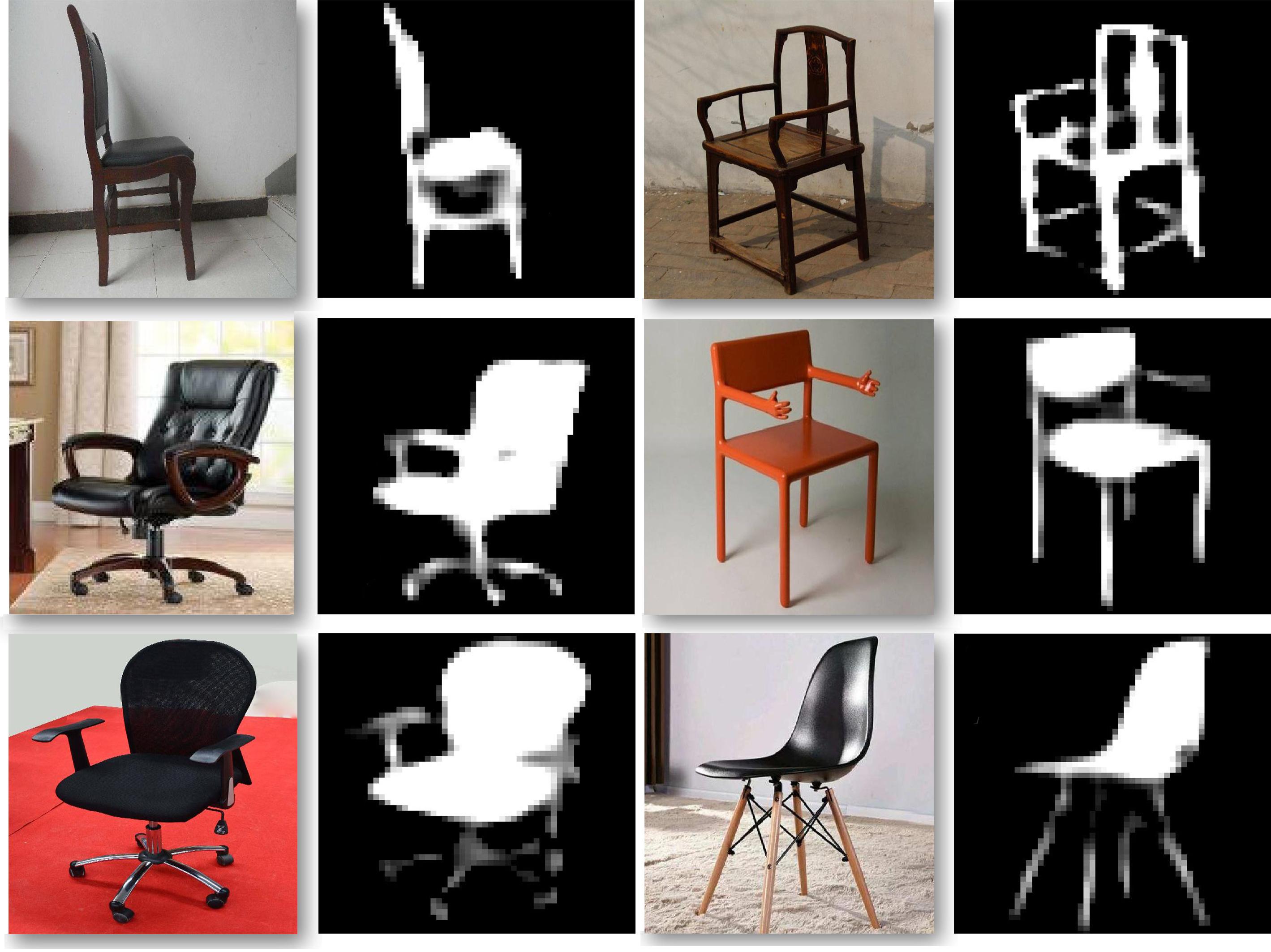}
\end{center}
   \caption{Object mask predictions for six sample images from the Internet. The probability of object mask is indicated by the brightness in the grey-scale output.}
\label{fig:result_mask}
\end{figure}

We first show in Fig.~\ref{fig:result_mask} some results of object mask prediction by our structure masking network.
As can be seen in the output, the background clutters are successfully filtered out and some detailed structures of the objects are captured.

\paragraph{Google image challenge for structure recovery.}
We first perform a qualitative evaluation on the capability and versatility of our structure recovery.
In order for a more objective study, instead of cherry-picking a few test images, we opt to conduct
a small-scale stress test with a Google image challenge~\cite{xu2011photo}.
During the test, we perform text-based image search on Google using the keywords of ``chair'', ``table''
and ``airplane'', respectively. For each search, we try to recover a 3D cuboid structure for each of the top 8 returned images using our method.

\begin{figure*}[t!]
    \centering
    \includegraphics[width=0.92\linewidth]{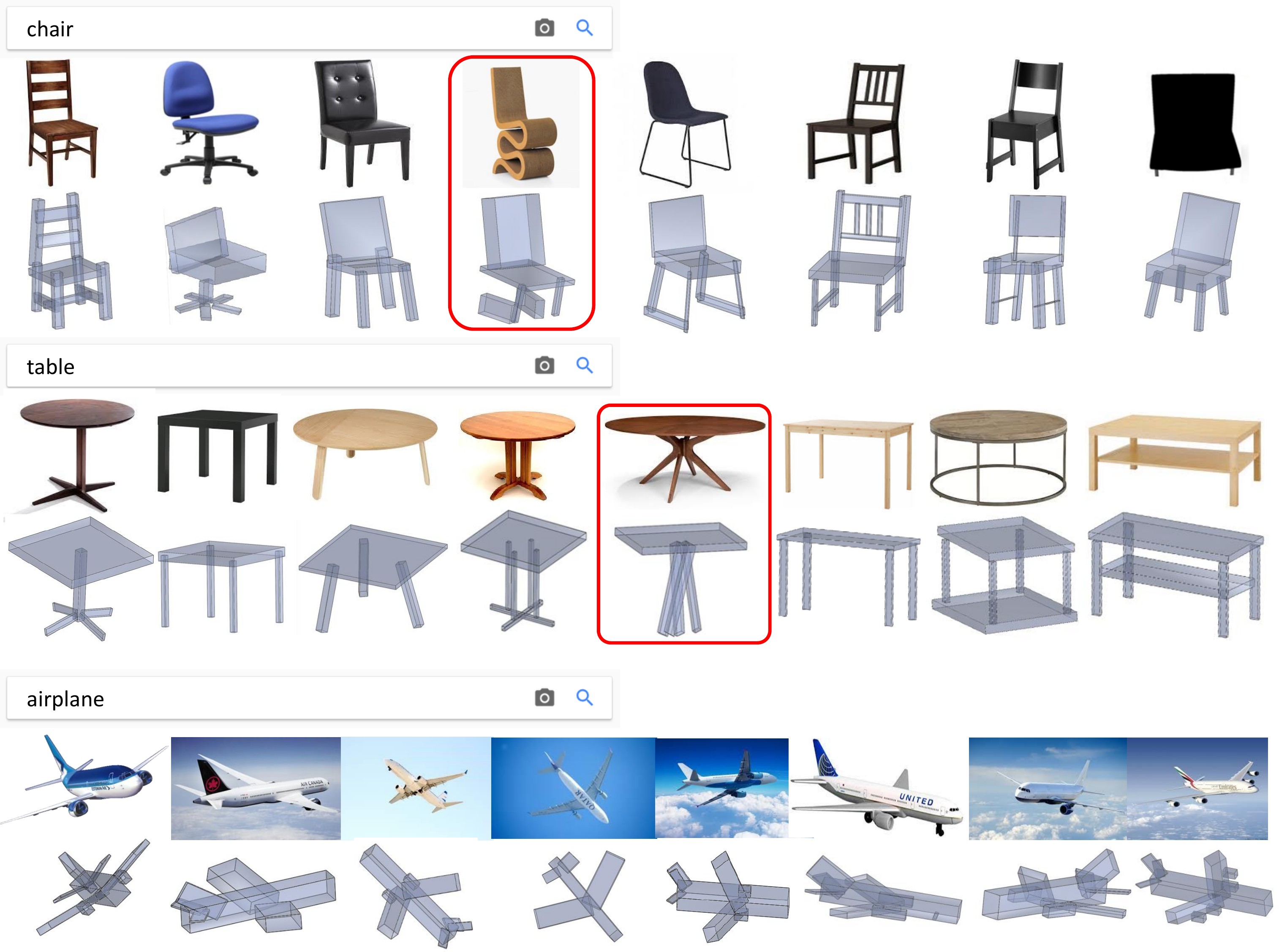}
    \caption{Google image search challenge for 3D shape structure recovery. We perform text-based image search on Google with keywords ``chair'', ``table'' and ``airplane'', respectively. For each search, we run our method on the top 8 returned images (duplicate or very similar images are removed). Failure cases are marked with red boxes.}
    \label{fig:result_internet}
\end{figure*}

The results are shown in Fig.~\ref{fig:result_internet}.
From the results, we can see that our method is able to recover 3D shape structures from real images in a detailed and accurate way. More importantly, our method can recover the connection and symmetry relations of the shape parts from single view inputs, leading to high quality results with coherent and plausible structure. Examples of symmetry recovery include the reflectional symmetry of chair legs or airplane wings, the rotational symmetry of legs in a swivel chair or a table.

There are some failure cases (marked with red boxes in Fig.~\ref{fig:result_internet}).
The marked chair example is not even composed of multiple parts and hence may not admit a part structure.
When the structure of the object of interest is unseen from our training dataset of 3D shapes, such as the marked table example, our method fails to recover a reasonable structure.

\paragraph{Quantitative evaluation.}
We quantitatively evaluate our algorithm with our test dataset.
For the structure masking network, we evaluate the mask accuracy by the overall pixel accuracy
and per-class accuracy against the ground-truth mask (see table~\ref{table:masknetresult}).
We provide a simple ablation study by comparing our method with two baselines:
single-scale (without refinement network) and two-scale (without jump connection).
The results demonstrate the effectiveness of our multi-scale masking network.

\begin{table}[h]\center
    \scalebox{0.95}{	
		\begin{tabular}{| c | c | c | }
			\hline
			Method & Overall Pixel & Per-Class \\ \hline
			single-scale & 0.953 &  0.917 \\ \hline		
            two-scale (w/o jump) & 0.982 &  0.964 \\ \hline
            two-scale (with jump) & 0.988 &  0.983 \\ \hline	
		\end{tabular}
    }
	\caption{An ablation study of structure masking network.}
    \label{table:masknetresult}
\end{table}

For 3D shape structure recovery, we develop two measures to evaluate the accuracy:

\begin{itemize}
\item \emph{Hausdorff Error}:  $ \frac{1}{2T}\sum_{i}^{T} (D(S_i, S^\text{gt}_i) +D(S^\text{gt}_i, S_i))$,
where $S_i$ is a recovered shape structure (represented by a set of boxes) and $S^{\text{gt}}_i$ its corresponding ground-truth. $T$ is the number of models in the test dataset.
$D(S_1, S_2) = \frac{1}{n}\sum_{B^1_j\in{S_1}}\min\limits_{B^2_k\in{S_2}}H(B^1_j, B^2_k)$ measures the
averaged minimum Hausdorff distance from the boxes in structure $S_1$ to those in $S_2$, where
$B^1_j$ and $B^2_k$ represent the boxes in $S_1$ and $S_2$, respectively.
$H(B^1,B^2) = \max\limits_{p\in{B^1}}\min\limits_{q\in B^2} ||p-q||$ is the Hausdorff
distance between two boxes, with $p$ and $q$ being the corner points of box.
Since Hausdorff is asymmetric, the distance is computed for both directions and averaged.


\item \emph{Thresholded Accuracy}: The percentage of boxes $B_{i}$ such that $\delta = H(B_i, B^{*}_i)/L(B^{*}_i) < threshold $, where $B_i$ is the $i$-th box in recovered shape structure $S$ and $B^{*}_i$ its nearest box in the ground-truth $S^\text{gt}$. $H$ is the Hausdorff distance between two boxes as defined above. $L$ is the diagonal length of a box.

\end{itemize}

We consider as our baseline where the structure masking network is simply a vanilla VGG-16 network. In table~\ref{table:comparison}, we compare the accuracy of structure recovery, based on the above two measures, for our method and the baseline. We also compare the two methods where VGG-16 is replaced with VGG-19.
The results demonstrate the significant effect of our structure masking network in helping the structure decoding.
This can also be observed from the reconstruction error plotted in Figure~\ref{fig:Loss} (bottom).
A deeper structure masking network (with VGG-19) also boosts the performance to a certain degree.

\begin{table}[h]\center
    \scalebox{0.9}{	
		\begin{tabular}{| c | c | c | c | }
			\hline
			\multirow{2}{*}{Method} & \multicolumn{1}{c|}{Hausdorff} & \multicolumn{2}{c|}{Thresholded Acc.} \\ \cline{3-4}
            \multicolumn{1}{|r|}{}  & \multicolumn{1}{c|}{Error} &  $\delta<0.2$ & $\delta<0.1$ \\ \hline
			Vanilla VGG-16 & 0.0980 & $ 96.8\% $ & $ 67.8\% $ \\ \hline
			Structure masking (VGG-16) & 0.0894 & $ 97.8\% $ & $ 75.3\% $ \\ \hline
			Vanilla VGG-19 & 0.0922 & $ 96.4\% $ & $ 72.2\% $  \\ \hline
			Structure masking (VGG-19) & 0.0846 & $ 97.6\% $ & $ 78.5\% $  \\ \hline
		\end{tabular}
    }
	\caption{Comparison of structure recovery accuracy over different methods.}
		\label{table:comparison}
\end{table}

\begin{figure}[!t]
\begin{center}
\includegraphics[width=1.0\linewidth]{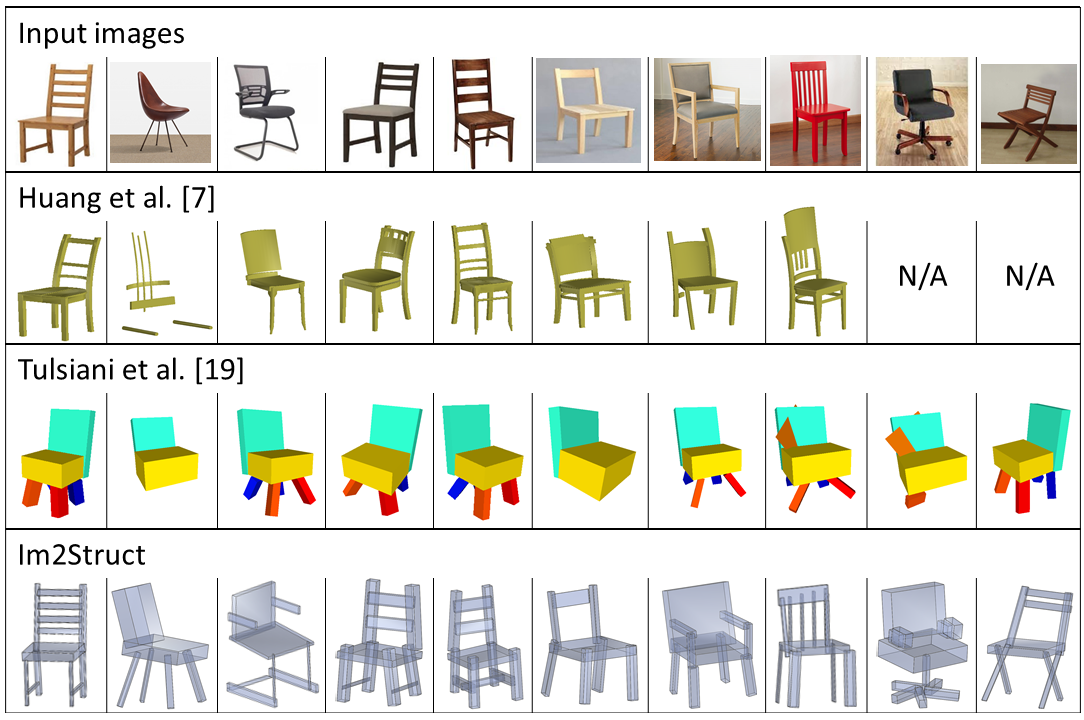}
\end{center}\vspace{-10pt}
   \caption{Comparing single-view, part-based 3D shape reconstruction between our Im2Struct and two alternatives.}
\label{fig:compare_star}
\end{figure}

\paragraph{Comparison.}
In Fig.~\ref{fig:compare_star}, we give a visual comparison of 3D shape reconstruction
from single-view images between our method and two state-of-the-art methods, \cite{huang2015single} and \cite{tulsiani2016learning}.
Both the two alternatives produce part-based representation of 3D shapes, making them comparable to our method.
The method by Huang et al.~\cite{huang2015single} recovers 3D shapes through assembling parts from database
shapes while preserving their symmetry relations.
The method of Tulsiani et al.~\cite{tulsiani2016learning} generates cuboid representation similar to ours,
but does not produce symmetry relations.
As can be seen, our method produces part structures which are more faithful to the input, due to the integration of the structure masking network, and meanwhile structurally more plausible, benefiting from our part relation recovery.




\vspace{-10pt}

\section{Applications}

We develop two prototype applications to demonstrate the utility of the recovered shape structure in structure-aware shape editing and processing.

\begin{figure}[!t]
\begin{center}
\includegraphics[width=\linewidth]{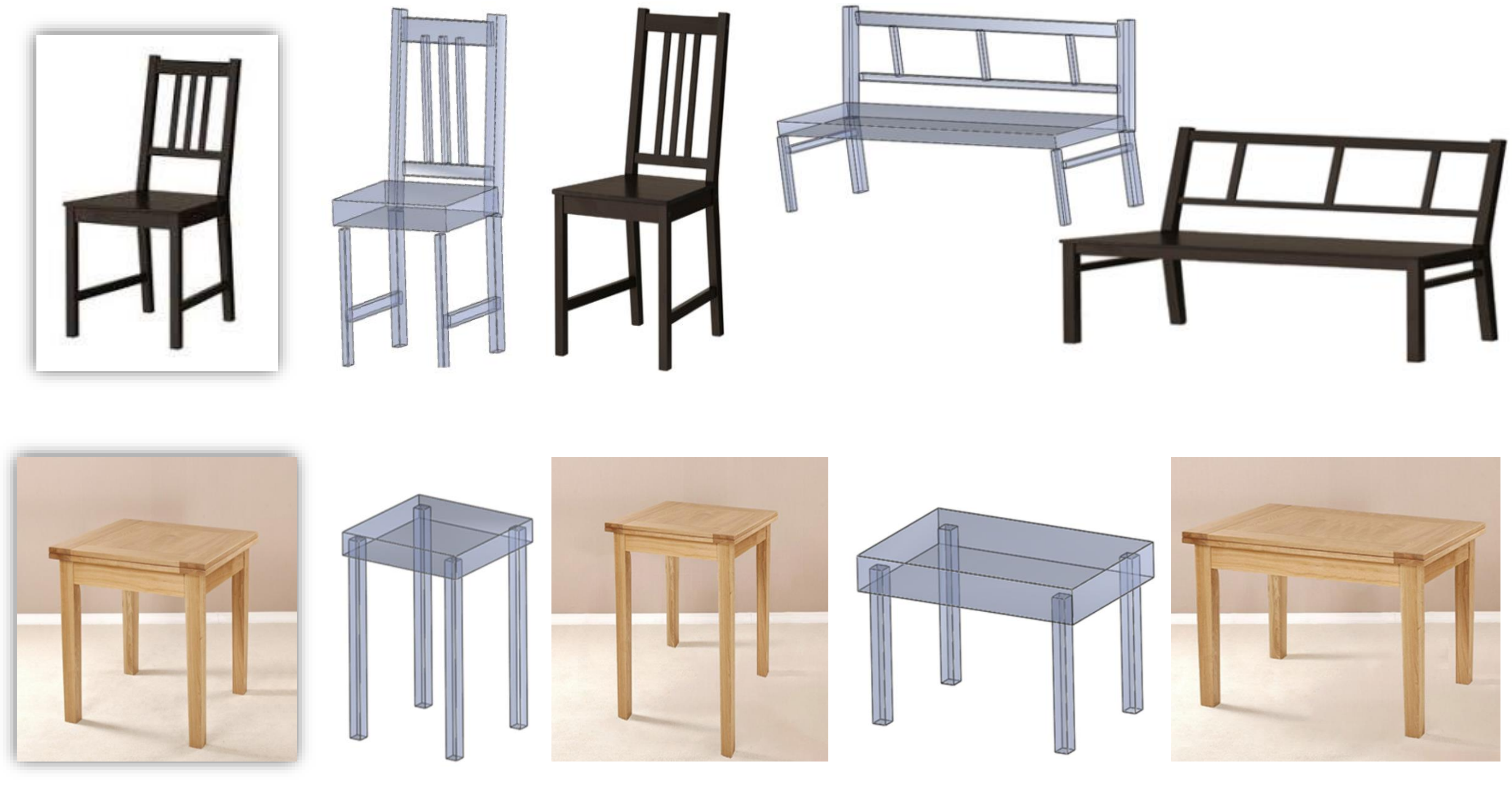}
\end{center}\vspace{-10pt}
   \caption{Examples of structure-aware image editing. Given an input image (the left column), we show the edited structures of 3D cuboids along with the edited images.}\vspace{-15pt}
\label{fig:application_image}
\end{figure}

\vspace{-15pt}
\paragraph{Structure-aware image editing.}
In~\cite{zheng_sigg12}, a structure-aware image editing is proposed, where a cuboid structure is
manually created for the object in the image to assist a plausible shape deformation.
With our method, this cuboid structure can be automatically constructed. Moreover,
the part relations are also recovered which can be used to achieve structure-aware editing.
Given an RGB image, we first recover the 3D shape structure of the object of interest using our method.
To align the inferred cuboid structure with the input image,
we train another network to estimate the camera view.
The 3D cuboids are then projected to the image space according to the estimated view.
The object in the image is segmented with a CRF-based method constrained with the cuboid projections~\cite{xu2011photo}.
Each segment is assigned to a 3D cuboid based their image-space overlapping. At this point, the image editing method in~\cite{zheng_sigg12} can be employed to deform the object of interest.
Fig.~\ref{fig:teaser} and~\ref{fig:application_image} show a few examples of structure-aware image editing based on our 3D shape structure recovery.

\begin{figure}[t]
\begin{center}
\includegraphics[width=1.0\linewidth]{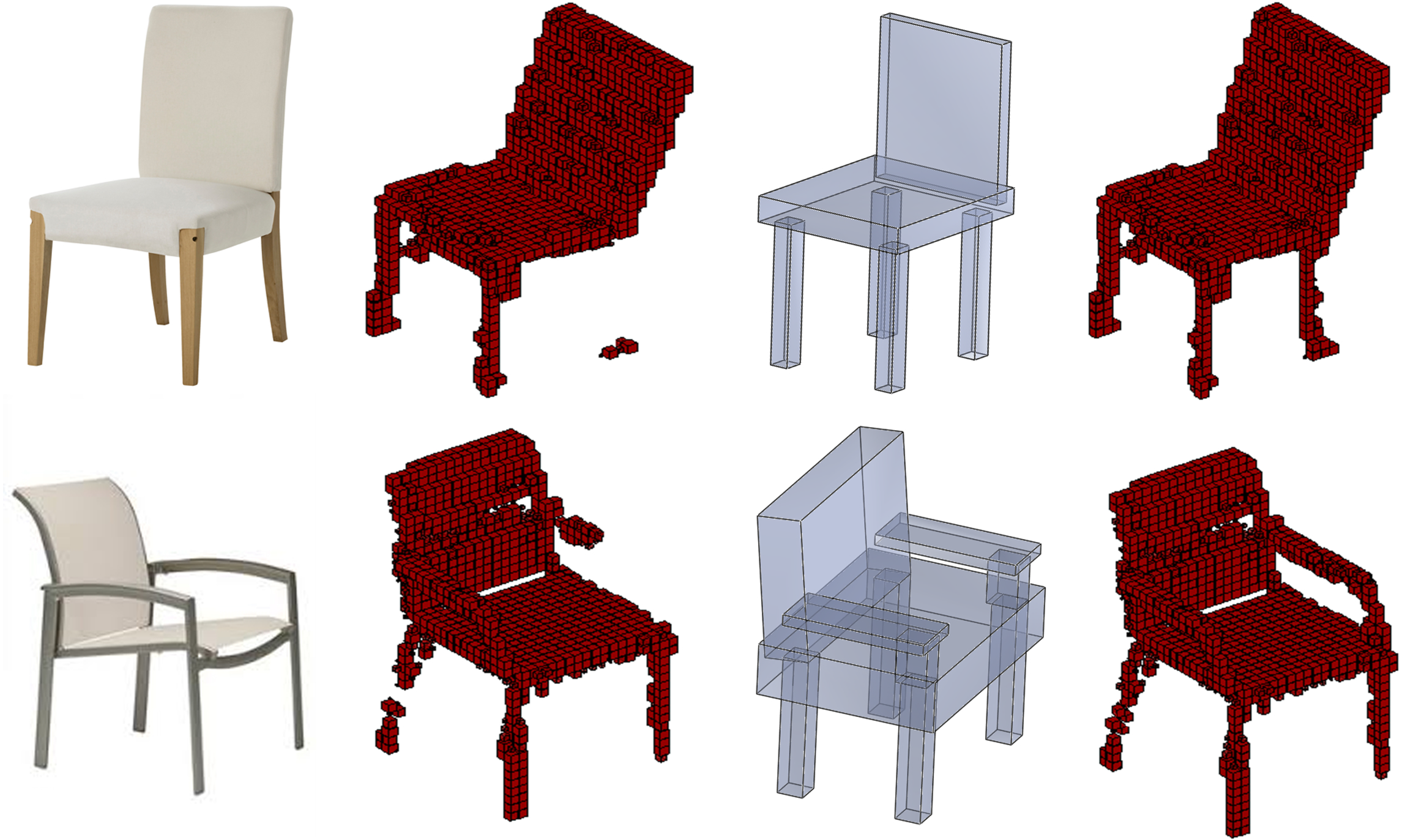}
\end{center}\vspace{-10pt}
   \caption{Part symmetry induced volume refinement. Given a 2D image, a volumetric shape is recovered using 3D-GAN~\cite{wu2016learning} where the missing voxels break the symmetry of the object. Our recovered 3D structure helps complete the volume based on part symmetry relation.}\vspace{-15pt}
\label{fig:application_voxel}
\end{figure}

\paragraph{Structure-assisted 3D volume refinement.}
A common issue with 3D reconstruction with volumetric shape representation is that the resolution of volume
is greatly limited due to the high computational cost. This usually results in missing parts
and hence broken structure in the reconstructed volume.
Our recovered 3D structures can be used to refine the 3D volumes estimated by existing approaches such as 3D-GAN~\cite{wu2016learning}.
Given a 2D image, a 3D volume is estimated with 3D-GAN and a cuboid structure recovered by our method.
They can be easily aligned with the help of camera view estimation (as have been done above).
Each voxel is assigned to the closest cuboid, leading to a part-based segmentation of the volume.
We then utilize the part symmetry relation in our recovered structure to complete the missing voxels; see results in Fig.~\ref{fig:application_voxel}.

\section{Conclusion}

We have proposed a deep learning framework that directly recovers 3D shape structures from single 2D images.
Our network joins a structure masking network for decerning the object structure and a structure recovery network for inferring 3D cuboid structure. The recovered 3D structures achieve both fidelity with respect to the input image and plausibility as a 3D shape structure. To the best of our knowledge, our work is the first that recovers detailed 3D shape structures from single 2D images.

Our method fails to recover structures for object categories unseen from the training set. For such cases, it would be interesting to learn an incremental part assembler. Our method currently recovers 3D cuboids only but not the underlying part geometry. A worthy direction is to synthesize detailed part geometry matching the visual appearance of the input image. Another interesting topic is to study the profound correlation between 2D features and 3D structure, so as to achieve a more explainable 3D structure decoding.






{\small
\bibliographystyle{ieee}
\bibliography{im2struct}
}

\end{document}